\documentclass[preprint,12pt]{elsarticle}

\usepackage{amssymb}

\usepackage{booktabs}
\usepackage{amsmath}
\usepackage{algorithm}
\usepackage{algpseudocode}
\usepackage{subcaption} 
\usepackage{hyperref}   
\hypersetup{
  colorlinks=true,
  citecolor=blue,
  linkcolor=blue,
 urlcolor=blue}
\usepackage{varwidth}

\usepackage{graphicx}

\newtheorem{mydef}{Definition}

\begin{document}

\begin{frontmatter}



\title{Semi-Supervised Hierarchical Multi-Label Classifier Based on Local Information}


\author[l1]{Jonathan Serrano-Pérez}
\ead{js.perez@inaoep.mx}
\author[l1]{L. Enrique Sucar}
\ead{esucar@inaoep.mx}

\affiliation[l1]{organization={Computer Science Dept., Instituto Nacional de Astrofísica, Óptica y Electrónica},
            city={San Andres Cholula},
            postcode={72840}, 
            state={Puebla},
            country={México}}

\begin{abstract}
Scarcity of labeled data is a common problem in supervised classification, since hand-labeling can be time consuming, expensive or hard to label; on the other hand, large amounts of unlabeled information can be found.
The problem of scarcity of labeled data is even more notorious in hierarchical classification, because the data of a node is split among its children, which results in few instances associated to the deepest nodes of the hierarchy.
In this work it is proposed the \textit{semi-supervised hierarchical multi-label classifier based on local information} (SSHMC-BLI) which can be trained with labeled and unlabeled data to perform hierarchical classification tasks.
The method can be applied to any type of hierarchical problem, here we focus on the most difficult case: hierarchies of DAG type, where the instances can be associated to multiple paths of labels which can finish in an internal node.
SSHMC-BLI builds pseudo-labels for each unlabeled instance from the paths of labels of its labeled neighbors, while it considers whether the unlabeled instance is similar to its neighbors. Experiments on 12 challenging datasets from  functional genomics show that making use of unlabeled along with labeled data can help to improve the performance of a supervised hierarchical classifier trained only on labeled data, even with statistical significance.

\end{abstract}




\end{frontmatter}


\section{Introduction}\label{sec1}
Scarce data is a common problem in supervised classification, this occurs when hand-labeling data is time-consuming, expensive or difficult to label \citep{Engelen2019-aSoSSL}. Consequently, training a classifier with few labeled data could produce an unreliable classifier.
In the same way, the problem of scarcity of labeled data may be found in a scenario of multi-label classification, that is, when instances are associated to multiple labels. 
Even more, this problem can be found in a hierarchical classification scenario where the instances are associated to not only a single path of labels but multiple paths.

In hierarchical classification the labels are arranged in a predefined structure (a hierarchy) which commonly is a tree but in its general form is a directed acyclic graph. The hierarchy contains the relations among the labels; and the instances can be associated to multiple labels, labels that form a path or multiple paths with respect to the hierarchy. These characteristics make hand-labeling more difficult and time-consuming for creating hierarchical datasets than for binary or multi-class datasets. 
Furthermore, the problem of scarcity of labeled data is even more notorious in hierarchical classification, because the data of a node is split among its children, which results in few instances associated to the deepest nodes of the hierarchy. 

Nonetheless, large amounts of data can be obtained from different sources of information, such as the Internet. Text, images, videos, etc., is data commonly desired for training a wide variety of classifiers, however, most of that information is unlabeled. Moreover, as previously discussed, the data could be required in scenarios where instances are associated to multiple labels, like hierarchical classification\citep{metzFreitas2009-ehcssl}, which makes more challenging to make use of unlabeled data. So suitable semi-supervised hierarchical classifiers that take advantage of unlabeled data are requiered.

In this way, the question that guides this research is:
\textit{will training a semi supervised hierarchical classifier with unlabeled and labeled data produce a classifier with better performance than a hierarchical classifier trained only on labeled data?}. This in a scenario where the hierarchy is any directed acyclic graph (DAG), and the instances can be associated to multiple paths of labels which can finish in an internal node.

Few works have been proposed for semi-supervised hierarchical classification \cite{metzFreitas2009-ehcssl,santosCanuto2014-ASSLHMLC,santosCanuto2014-ASTSSLHMLM,xiao2019-EPPSSWSHTC,levatic-2022-sslpct,Serrano-Sucar-2022-SSHCBLI};  most are proposed for hierarchies of tree type, being unable to handle hierarchies where a node has more than one parent; moreover, none of them carried out a study for hierarchies of DAG type (hierarchies that allow nodes to have more than one parent) with instances associated to multiple paths of labels in an inductive way.



Therefore, \textit{semi-supervised hierarchical multi-label classifier based on local information} (SSHMC-BLI\footnote{Link: \url{https://github.com/jona2510/SSHMC-BLI/tree/master} }) is the proposed method which can handle hierarchies of DAG type and it is designed for instances that are associated to multiple paths of labels. Its main idea is to pseudo-label the unlabeled data, which are later used to train a hierarchical multi-label classifier.
It builds pseudo-labels using the labels of the nearest labeled neighbors to each unlabeled instance, then, the function \textit{similarity of an instance with a set of instances} (SISI) \citep{Serrano-Sucar-2022-SSHCBLI} is used to determine if the unlabeled instance is similar to its labeled neighbors, if they are similar, the unlabeled instance is pseudo-labeled, else it stays unlabeled. Experiments on the Gene Ontology (GO) collection were carried out, where the performance of the proposed method was compared against its supervised counterpart.
The proposed method shows outstanding performance; it outperformed the results obtained by its baseline, a supervised hierarchical classifier trained only on labeled instances, while considering different amounts of labeled and unlabeled data.

The main contributions of this manuscript\footnote{A preliminary version of this work was published in IBERAMIA-2022 \citep{Serrano-Sucar-2022-SSHCBLI}. In this work, the method is extended to handle hierarchies of directed acyclic graph instead of only trees, also the instances can be associated to multiple paths of labels instead of only one in the aforementioned paper. Experiments on Gene Ontology datasets were carried out where the hierarchies are DAGs. Furthermore, a statistical analysis was carried out for comparison of the proposed and standard methods over multiple datasets.} are: (i) a semi-supervised hierarchical multi-label classifier that can handle hierarchies of directed acyclic graph type and it is designed for instances that are associated to multiple paths of labels; (ii) an experimental comparison of the proposed method on several real world datasets against its supervised counterpart and standard methods.

The document is organized as follow. Section \ref{s:F} summarizes fundamentals of hierarchical classification and semi-supervised learning. Section \ref{s:sshc} presents semi-supervised hierarchical classification. Section \ref{s:RW} reviews related work. Section \ref{s:PM} presents the proposed method. Sections \ref{s:datasets} introduces the datasets that are used in the experiments. Section \ref{s:ER} presents the experiments and results.  Finally, in section \ref{s:CFW}, conclusions and some ideas for future work are given.

\section{Fundamentals}\label{s:F}
\subsection{Hierarchical classification}
Hierarchical classification (HC) can be seen as a special type of multi-label classification, in which the labels are arranged in a predefined structure which is a tree or in its general form a directed acyclic graph (DAG). The hierarchy or \textit{hierarchical structure} ($HS$) can be denoted with graph notation: $HS=(L,E)$, where $L$ is the set of labels/nodes and  $E$ is the set of edges that links the nodes. Finally, $HS$ is a DAG.

Furthermore, the subset of labels for an instance has to comply the \textit{hierarchical constraint}. The hierarchical constraint states that if an instance $x$ is associated to the label $l \in L$ then $x$ has to be associated to the ancestors of $l$, $Anc(l)$, given by the HS:
\begin{equation}
    \forall x \in l \rightarrow  x \in z, \forall z \in Anc(l)    
\end{equation}
Therefore, a \textit{valid} or \textit{consistent path} is a subset of the labels that complies the hierarchical constraint. 

Therefore, the problem of hierarchical classification consist in assigning to a particular object described by $d$ attributes, a subset of labels that comply the hierarchical constraint: $f_{HC}=\mathbb{R}^{d} \rightarrow  \{0,1\}^{\mid L \mid }$.

\subsubsection{Hierarchical Classification Problems} \label{s:hcproblems}
In hierarchical classification there are several type of problems as shown by Silla and Freitas \cite{Silla2011}. They describe the hierarchical problems with 3 aspects $<\Upsilon ,\Psi ,\Phi>$, where:

\begin{itemize}
    \item $\Upsilon$: specifies the type of hierarchical structure in which the labels are arranged, so, it can take one of two values, $T$ if it is a tree or $DAG$ if it is a direct acyclic graph.
    \item $\Psi$: specifies whether an instance can be associated with either one or multiple paths. Thus the values that it can take are $SPL$: single path of labels,  and $MPL$: multiple paths of labels.
    \item $\Phi$: describes the depth of the paths of the instances, two values are allowed $FD$: full depth, if the paths of all the instances reach a leaf node; and $PD$: partial depth, if at least one path of an instance does not reach a leaf node.
\end{itemize}
Therefore, there are eight different types of hierarchical classification problems, Table \ref{table_hcp} lists them. It is important to know the hierarchical classification problem in order to choose the most suitable method, since each method tends to perform better in a specific type of problem(s).

\begin{table}[tb]
\caption{List of the different hierarchical classification problems. $\Upsilon$: hierarchical structure, $\Psi$: number of paths, $\Phi$: depth of paths.}
\centering
\label{table_hcp}
\begin{tabular}{ccc}
\hline
\textbf{$\Upsilon$} & \textbf{$\Psi$} & \textbf{$\Phi$} \\ \hline
T                   & SPL             & PD              \\
T                   & SPL             & FD              \\
T                   & MPL             & PD              \\
T                   & MPL             & FD              \\
DAG                 & SPL             & PD              \\
DAG                 & SPL             & FD              \\
DAG                 & MPL             & PD              \\
DAG                 & MPL             & FD              \\ \hline
\end{tabular}
\end{table}

The main approaches for hierarchical classification \citep{naik-bookHC-2018,Silla2011} are briefly described next:
\begin{itemize}
    %
    
    \item \textit{Local Classifier per Node (LCN)}: For each node of the hierarchy, except the root node, a binary classifier is trained, which is commonly used to predict whether an instance is associated to the node.

    \item \textit{Local Classifier per Parent Node (LCPN)}: In  this approach, for each non-leaf node (including the root node) a multiclass classifier is trained to predict its children nodes. 
    
    \item \textit{Local Classifier per Level (LCL)}: The LCL approach consists in training a multi-class classifier for each level of the hierarchy. This approach is less used than the previous two approaches.
    
    \item \textit{Global Classifier (GC)}: They consider the entire class hierarchy at once.
    
    \item \textit{Flat}: Even though these methods may not be called \textit{hierarchical} classifiers, in hierarchical classification the methods that ignore the hierarchy and focus their training and predictions only on the leaf nodes of the hierarchy belong to the \textit{flat} approach.
    
\end{itemize}{}

\subsubsection{Local Classifier per Node} \label{s:TD}
LCN are used in this work for training the semi-supervised classifier.

In the training phase a \textit{policy} is defined in order to select the positives and negatives instances for each binary classifier. Some policies have been proposed \citep{Eisner-policies,fagni-policies}, however, new policies or variants  can be proposed, some of them are described next, let $l \in L$:

\begin{itemize}
    \item \textit{Less inclusive policy}: for a node $l$, the positive instances are all the instances associated to $l$, and the negative instances are the rest.
    \item \textit{Inclusive policy}: for each node $l$, the positive instances are all the instances associated to $l$, while the negatives are the rest except instances that are associated to ancestors of $l$.
    \item \textit{Siblings policy}: for each node $l$, the positive instances are all the instances associated to $l$, while the negatives are those associated to the siblings of $l$.
    \item \textit{Exclusive policy}: for each node $l$ the positive instances are only those instances which its most specific label is $l$, and the negatives are those instances which its most specific label is some sibling of $l$.
    \item \textit{Balanced bottom-up}: for each node $l$, the positive instances are all the instances associated to $l$, while the negatives are at most equal to the amount of positives, taking them first from its siblings, then from uncles and so on.
\end{itemize}
The proposed method, SSHMC-BLI, makes use of the balance bottom-up policy.

\subsection{Semi-Supervised Learning}
Semi-Supervised Learning (SSL) can be seen as the branch of machine learning that combines supervised and unsupervised learning \citep{chapelle2010-SSLbook,zhu2008-sslls}; that is, SSL algorithms can handle labeled and unlabeled data to perform learning tasks.
Semi-supervised classification methods are appropriate in scenarios where labeled data is scarce and unlabeled data is available. Scarce labeled data occurs because it is expensive or difficult to obtain, for instance in computer-aided diagnosis, drug discovery and part-of-speech tagging \citep{Engelen2019-aSoSSL}.

\begin{figure}[tb]
	\centering
    \includegraphics[width=0.9\columnwidth]{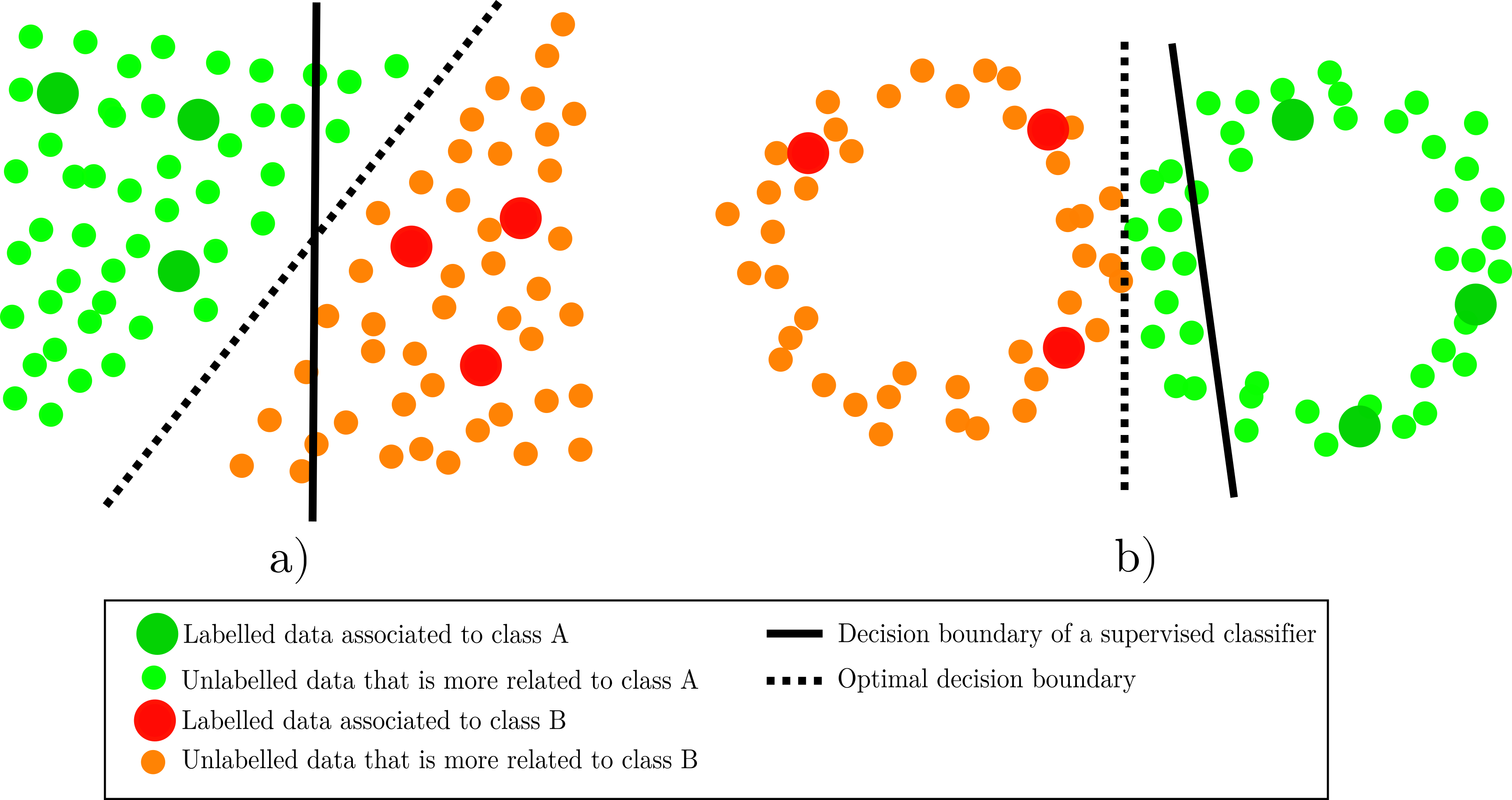}
    \caption[Semi-Supervised Learning Assumptions.]{Semi-supervised learning assumptions \citep{Engelen2019-aSoSSL}. a) Smoothness and low-density assumptions, b) Manifold assumption. (Best seen in color.)}
  \label{ssl-assumptions}
\end{figure}

In SSL there are some assumptions on which most semi-supervised learning algorithms are based on, so they intent to satisfy one or more of them \citep{chapelle2010-SSLbook}. They are briefly described below:
\begin{itemize}
    \item \textit{Smoothness assumption}: It says that for two input points $x_{i}, x_{j} \in X$ which are close by in the input space, their labels $y_{i}, y_{j}$ should be the same. Furthermore, this assumption can be applied transitively to unlabeled data.
    
    \item \textit{Low-density assumption}: It states that the decision boundary of a classifier should not pass through high-density areas in the input space. That is, the decision boundary  should preferably pass through low-density regions.
    
    \item \textit{Manifold assumption}: this one is similar to the smoothness assumption, that is, it says that the input space is composed of multiple lower dimension manifolds on which all data points lie, therefore, data points on the same manifold must have the same label.
\end{itemize}
Examples of smoothness, low-density and manifold assumptions are shown in Fig. \ref{ssl-assumptions}. As it can be seen, the decision boundaries obtained by a supervised classifier are not optimal because they are learned with only labeled data; on the other hand, the semi-supervised methods take advantage of labeled and unlabeled data to learn a decision boundary close to the optimal. The proposed method in this work is based on the smoothness assumption.

Semi supervised learning methods are commonly divided into two main groups, inductive and transductive \citep{Engelen2019-aSoSSL,zhu-2009-ISSL}. The former produce a classification model that is used to predict the labels of instances that were not used in training, while the second is focus only on labeling the unlabeled instances. In this way, the proposed method belongs to the inductive group, because a classification model for predicting new data is generated. 

\section{Semi-Supervised Hierarchical Classification} \label{s:sshc}
Formally, we define semi-supervised hierarchical classification as a tuple $SSHC =\; <HS,(X,Y),U>$, where: 
\begin{itemize}
    \item $HS = (L,E)$ is a \textit{directed acyclic graph} that represents the hierarchy, where $L$ is the set of nodes and $E$ is the set of edges that link the nodes.
    \item $(X,Y)$ is the labeled set. $X = \{x_{1},x_{2},...,x_{n}\}$ contains $n$ instances, where $x_{i} \in \mathbb{R}^{d}$, that is, each instance $x_{i}$ is described by a vector of $d$ attributes. Let $Y = \{y_{1},y_{2},...,y_{n}\}$ be the labels for $X$, where $y_{i} \in \{0,1\}^{\mid L \mid}$, that is, each $y_{i,j}$ indicates whether the $i$-th instance is associated to the $j$-th label, while $y_{i}$ satisfies the \textit{hierarchical constraint}.
    \item $U$ is the unlabeled set. $U = \{x_{n+1},x_{n+2},...,x_{n+m}\}$ contains $m$ instances described by the same $d$ attributes as in $X$.
\end{itemize}
SSHC is composed by three main elements, a hierarchy, a labeled set and a unlabeled set.

Hence, the problem of \textit{semi-supervised hierarchical classification} consist in assigning to a particular object described by $d$ attributes, a subset of labels that comply the hierarchical constraint:
\begin{equation}
    f_{SSHC}: \mathbb{R}^{d} \rightarrow  \{0,1\}^{\mid L \mid}
\end{equation}
However, commonly in hierarchical classification where the instances are associated to multiple paths of labels \citep{vens2008decision,Cerri-2016-RSHMLCPFP,giunchiglia2020-CHMLCN}, the problem is \textit{modified} to assign to a particular instance, the probability of being associated to each node:
\begin{equation}
    f_{SSHC}: \mathbb{R}^{d} \rightarrow  \{[0,1]\}^{\mid L \mid}
\end{equation}
However, this prediction has to comply the \textit{hierarchical probability constraint}, which is defined next:

\begin{mydef}
    \textbf{Hierarchical probability constraint} states that the probability for an instance in the node $l$ has to be equal or lower than the probabilities of all the parent nodes of the node $l$; let $f$ be a model with one output per node, then: 
    \begin{equation*}
        f_{l} \leq f_{z}, \; \forall z \in Parents(l);\; \forall l \in L 
    \end{equation*}
\end{mydef}
Even though, the proposed method is general enough to be applied to any of the hierarchical problems described in Section \ref{s:hcproblems}, this work will be focused on the most complex type of the hierarchical problems, the one described as $<DAG,MPL,PD>$, that is, where the hierarchy is a directed acyclic graph  and the instances are associated to multiple paths of labels, paths that could finish in internal nodes of the hierarchy.
Hence, it is hypothesized that training a model with labeled and unlabeled data in this type of hierarchical problems will perform better than training a hierarchical classifier only on labeled data:

\begin{equation}
    performance(f_{SSHC}) \geq performance(f_{HC})
\end{equation}


\subsection{Related Work}\label{s:RW}
This section presents the works that addressed semi-supervised learning for hierarchical classification and shows how the proposed method is different/related from the rest.

Initially, two standard methods are presented. First, \textit{self-training for multi-label classification} (STML) which for each label a binary classifier is self-trained using the whole unlabeled set in each; the prediction for a new instance is the union of the predictions of the binary classifiers; this method (as any multi-label classifier) ignores the hierarchy while training, and its predictions do not always comply the hierarchical constraint. Second, \textit{self-training  hierarchical classifier}\footnote{STHC is introduced in this paper as a \textit{standard} semi supervised classifier for hierarchical classification where the hierarchy is a DAG and the instances are associated to multiple paths of labels. Also, it can be seen as an extension of self-train A \citep{metzFreitas2009-ehcssl} but the predictions of the binary classifiers are post-processed to comply the hierarchical probability constraint instead of applying the top-down procedure. } (STHC) where each node of the hierarchy is self-trained like STML, but in the prediction phase, the predictions of the local classifiers are post-processed such that they comply the hierarchical probability constraint, that is, if the probability in a node is greater than the probability of its parent with the lowest probability, its probability is reduced down to the probability of that parent.

The first method for semi-supervised hierarchical classification was proposed by Metz and Freitas \cite{metzFreitas2009-ehcssl}. It is a Top-Down classifier that can handle hierarchies of tree type and predicts a single path of labels that always reaches a leaf node. Decision trees are trained as binary classifiers for each node, then each classifier is self-trained following one of three strategies: \textit{self train A, B} and \textit{C}. Nevertheless, the reported results are not superior in a statistically significant way against the supervised hierarchical classifier.  

Hierarchical multi-label classification using semi-supervised label powerset (HMC-SSLP) was proposed by Santos and Canuto \cite{santosCanuto2014-ASSLHMLC}. It consist on training a hierarchical multi-label classification with label powerset (HMC-LP) \citep{cerriCarvalo2009-CLGHMLCMUDT} with labeled data, then it is used to pseudo-label a predefined proportion of the unlabeled data which are added to the training set for the next iteration, this process iterates until all unlabeled data is pseudo-labeled.  HMC-LP combines all the classes of each example to generate a new hierarchy, nevertheless, examples of how to combine paths of different lengths are not shown. 

Hierarchical multi-label using semi-supervised random k-labelsets (HMC-SSRAkEL) by Santos and Canuto \cite{santosCanuto2014-ASSLHMLC}. This method trains LCPN's, that is, for each non-leaf node a RAkEL classifier \citep{tsoumakas2011-RAkEL} is trained to predict its children. Later, a Top-Down procedure is used to pseudo-label a predefined proportion of the unlabeled data, which are added to the training set for the next iteration, this process iterates until all the unlabeled data is pseudo-labeled. Nevertheless, this method lacks of a way to select the instances with the most confident predictions, instead, it adds the whole set of pseudo-labeled instances to the labeled set in each iteration.

Santos and Canuto \cite{santosCanuto2014-ASTSSLHMLM} proposed hierarchical multi-label classification using semi-supervised binary relevance (HMC-SSBR) which is the semi-supervised version of HMC-BR \citep{cerriCarvalo2009-CLGHMLCMUDT}. Santos and Canuto indicate that BR is replaced by SSBR \citep{santosCanuto2012-USSLMLCP}, a semi-supervised method for \textit{multi-label classification}, that is, the unlabeled instances  are pseudo-labeled with the prediction of the TD, then the same steps than HMC-SSRAkEL to pseudo label the unlabeled instances are carried out; hence, HMC-SSBR also lacks of way to select the pseudo-labeled instances with the most confident predictions.


Path cost-sensitive algorithm with expectation-maximization (PCEM) was proposed by Xiao et al. \cite{xiao2019-EPPSSWSHTC} for hierarchical text classification. It consist of two main steps. First, path cost-sensitive naive Bayes classifier (PCNB) \citep{xiao2019-EPPSSWSHTC} is trained with the labeled data as the base classifier, then it is used to pseudo-label the unlabeled instances. Second, the PCNB is trained with labeled and pseudo-labeled instances, and this process is iterated until the parameters of the PCNB converge. 
Because PCNB is designed using the bag-of-words representation, it is not straightforward to apply PCEM in non-text domains.

Levatic et al. \cite{levatic-2022-sslpct} proposed semi-supervised predictive clustering trees (SSL-PCT), which 
is based on predictive clustering trees (PCT) \citep{breiman-1984-crt,blockeel-1998-tdict}. PCT's consist of a hierarchically organized set of clusters, where the root cluster is recursively divided into smaller cluster as one goes deeper to the leaves.
They reported that the results of SSL-PCT were \textit{not so successful} on the functional genomics datasets, because the supervised hierarchical classifier was rarely outperformed. In this work a transductive study was carried out,

Serrano-Pérez and Sucar \cite{Serrano-Sucar-2022-SSHCBLI} proposed semi-supervised hierarchical classifier based on local information (SSHC-BLI), which is based on the smoothness assumption. 
SSHC-BLI got better performance than the supervised classifier, showing that making use of unlabeled data can help to improve the performance of a hierarchical classifier trained only on labeled data. Nevertheless, SSHC-BLI only works for hierarchies of tree type, and the instances have to be associated to a single path of labels.

Finally, \textit{semi-supervised hierarchical multi-label classifier based o local information} (SSHMC-BLI) is the proposed method in this work,  which can be seen as a generalization of SSHC-BLI \citep{Serrano-Sucar-2022-SSHCBLI}, that is, SSHMC-BLI can handle any hierarchy of DAG type not only trees, and the instances can be associated to multiple paths of labels not only to a single. In this way, the method tries to pseudo-label each unlabeled instance making use of the labels of its labeled neighbors, while considering if the unlabeled instance is similar to its neighbors, if so, it can be pseudo-labeled; later a hierarchical multi-label classifier is trained with the labeled and pseudo-labeled instances. 

Table \ref{t:c_rw} compares the related and proposed methods. As it can be seen, the proposed method is the only one that can handle hierarchies of DAG type from the inductive methods. The proposed method  belongs to the \textit{inductive} approach, that is, the model trained with labeled and unlabeled data is used to predict the labels of instances that were not used in the training phase; in the opposite case of the transductive methods which are focused on pseudo-labeling the unlabeled set. 


\begin{table}[htb]
 \centering
 \caption{Comparative list of the related and proposed method. $\Upsilon$: hierarchical structure, $\Psi$: number of paths, $\Phi$: depth of paths, NA: not applicable.}
 \label{t:c_rw}
\begin{tabular}{@{}lcccc@{}}
\toprule
\textbf{Method}                                 & \textbf{$\Upsilon$} & \textbf{$\Psi$} & \textbf{$\Phi$} & \textbf{SSL approach} \\ \midrule
STML & NA                & NA             & NA              & Inductive             \\
STHC & DAG                & MPL             & PD              & Inductive             \\
Metz and Freitas \cite{metzFreitas2009-ehcssl} & Tree                & SPL             & FD              & Inductive             \\
HMC-SSLP \citep{santosCanuto2014-ASSLHMLC}       & Tree                & MPL             & PD              & Inductive             \\
HMC-SSRAkEL \citep{santosCanuto2014-ASSLHMLC}    & Tree                & MPL             & PD              & Inductive             \\
HMC-SSBR \citep{santosCanuto2014-ASTSSLHMLM}     & Tree                & MPL             & PD              & Inductive             \\
PCEM \citep{xiao2019-EPPSSWSHTC}                 & Tree                & SPL             & FD              & Inductive             \\
SSL-PCT \citep{levatic-2022-sslpct}             & DAG                 & MPL             & PD              & Transductive          \\
SSHC-BLI \citep{Serrano-Sucar-2022-SSHCBLI}     & Tree                & SPL             & FD              & Inductive             \\ \midrule
SSHML-BLI (proposed)                            & DAG                 & MPL             & PD              & Inductive             \\ \bottomrule
\end{tabular}
\end{table}


\subsection{Evaluation Measures}\label{s:EM}
The outputs of the proposed method are the probabilities for each node of the hierarchy. Hence, it is not suitable to use evaluation measures such as \textit{accuracy} or \textit{hierarchical F measure} \citep{Silla2011,Nakano-Pinto}, since they require binary values (associated, not associated) for each node. Even though, a threshold could be applied to the output of the model to get binary values, it is not straightforward to set the threshold, moreover, different thresholds could produce different results.

In this case, the evaluation measure \textit{area under the average precision and recall curve} $AU(\overline{PRC})$  also known as \textit{average precision} (AP) \citep{zhu-2004-RPAP} is used to evaluate the performance of the models:
\begin{equation}
    AP = \sum_{n}(R_{n}-R_{n-1})P_{n}
\end{equation}
where $P_{n}$ and $R_{n}$ are the precision and recall at the $n$-th threshold, respectively. Equations \ref{eq:pm}, \ref{eq:rm} correspond to precision and recall, respectively. 
\begin{equation}
    P = \frac{\sum_{i=1}^{\mid L \mid} TP_{i}} { \sum_{i=1}^{\mid L \mid}{TP_{i}} + \sum_{i=1}^{\mid L \mid}{FP_{i}}} \label{eq:pm}
\end{equation}
\begin{equation}
    R = \frac{\sum_{i=1}^{\mid L \mid} TP_{i}} { \sum_{i=1}^{\mid L \mid}{TP_{i}} + \sum_{i=1}^{\mid L \mid}{FN_{i}}} \label{eq:rm}
\end{equation}
AP is an evaluation measure independent of a threshold to determine whether an instance is associated to a node, which makes it ideal in this kind of scenarios. Furthermore, this measure has been used to evaluate the performance of several hierarchical multi-label methods \citep{Cerri-2016-RSHMLCPFP,CERRI-2014-HMLULNN,leander-2010-PGFUHMDTE}.

\section{Semi Supervised Hierarchical Multi-label Classifier Based on Local Information}\label{s:PM}

\begin{algorithm}[]
\caption{SSHMC-BLI algorithm}\label{alg:sshc-knn}
\begin{algorithmic}[1]
\Require $(X,Y)$: labeled data, $U$: unlabeled data, $k$: number of nearest labeled neighbors, $THR$: similitude threshold, $t2label$: threshold to pseudo-label an instance, $HS$: hierarchy of DAG type, $maxIterations$: maximum number of iterations.
\Ensure $f_{sshc}$: trained SSHMC-BLI classifier
\State $T \gets 1$  \Comment{Iteration}
\State $LD \gets X$ \Comment{LD: Labeled data}
\State $CL \gets L$ \Comment{Labels of labeled data}
\While{$True$}
	\For {\textbf{each} $u_{j} \in U$}
        \State $IND_{j} \gets getNLN(k,u_{j},LD)$ \Comment{Get the nearest labeled neighbors} \label{get-knn}
        \State $PSL_{j} \gets buildPseudoLabel(IND_{j},LD,t2label)$ \Comment{Pseudo label for $u_{j}$} \label{pseudo-label}
   	\EndFor

	\For {\textbf{each} $u_{j} \in U$ with valid $PSL_{j}$}
        \If {$SISI(u_{j},IND_{j})<THR$} \label{simil-u-ind}
            \State $PSL_{j}=\O$     \Comment{Invalid pseudo-label}
		\EndIf
   	\EndFor
        
    \If{$(T > maxIterations)$ or $(PSL^{T}==PSL^{T-1})$}
        \State \textbf{break}  loop (while)
    \Else \Comment{join labeled data with valid pseudo-labeled data}
        \State $CL \gets Y \cup valid(PSL)$ 
        \State $LD \gets X \cup U[valid(PSL)]$
    \EndIf
    \State $T \gets T+1$
\EndWhile
\State $f_{SSHC} \gets trainHMC(LD,CL,HS)$   \Comment{Train a hierarchical multi-label classifier}

\end{algorithmic}
\end{algorithm}

SSHMC-BLI  is based on the smoothness assumption, that is, neighboring instances must have the same or similar paths of labels.
SSHMC-BLI tries to build pseudo-labels for each unlabeled instance using  the paths of labels of its neighboring labeled instances, however, only if the unlabeled instance is similar to its labeled neighbors, it is pseudo-labeled, otherwise, it stays unlabeled; this process is iterated until all the pseudo-labels do not change.

The steps of SSHMC-BLI are shown in Algorithm \ref{alg:sshc-knn}. In general, it is an iterative method that tries to pseudo-labeled the unlabeled data using its nearest labeled neighbors as it is shown in lines \ref{get-knn} - \ref{pseudo-label}, the way in which pseudo-labels are built is described  in subsection \ref{s:pseudo-label}. This method considers the similitude of each unlabeled instances with its neighbors (line \ref{simil-u-ind}), such that if they are not \textit{similar} the unlabeled instance looses its pseudo-label; the function \textit{similitude of an instances with a set of instances} (SISI) \citep{Serrano-Sucar-2022-SSHCBLI} is used for estimating the similitude. Let $p$ be an instance, let $A$ be a set of instances, where $x_{i}\in A$, let $k$ be the length of A, and let $d(X,Y)$ be the euclidean distance, then, SISI is defined in equation \ref{eq:sisi}. 

\begin{eqnarray}
    SISI(p, A)=\left\{\begin{matrix}
        1 & uavg(p, A)<=lavg(A) \\ 
        0 & uavg(p, A)>=n*lavg(A) \\ 
        \frac{lavg(A)-uavg(p, A)}{(n-1)lavg( A)}+1 & otherwise
    \end{matrix}\right. \label{eq:sisi} 
\end{eqnarray}
\begin{eqnarray}
    lavg(A)=\frac{\sum_{i=1}^{k} \sum_{j=i+1}^{k} d(x_{i},x_{j})}{\frac{k(k-1)}{2}}\\
    uavg(p, A)=\frac{\sum_{i=1}^{k} d(p,x_{i})}{k}
\end{eqnarray}
In this way, SISI takes into account the distances among the labeled neighbors (lavg) and the distances of the unlabeled instance with its labeled neighbors (uvag). Furthermore it returns a score between $[0,1]$, where 1 indicates that the unlabeled instance is similar to its labeled neighbors, 0 otherwise.

Finally, the method finishes when the pseudo-labels for the unlabeled data do not change from an iteration to other, or when the maximum number of iterations is reached; so, with the labeled and pseudo-labeled data a hierarchical multi-label classifier (it is described in section \ref{s:base-sshmc-bli}) is trained.

Three variants of SSHMC-BLI method are presented in this work, the differences among them are described next: \textit{Variant 1} (V1), it follows Algorithm \ref{alg:sshc-knn} to the letter. \textit{Variant 2} (V2), in each iteration all pseudo-labels for the unlabeled set are re-built, so, after the first iteration an instance, that was added to the training set, will have to itself as one of its nearest labeled neighbors; in order to avoid this, the function \textit{getNLN} (line \ref{get-knn}) is modified to guarantee that none of the nearest labeled neighbors is the instance itself.  And \textit{Variant 3} (V3), taking into account that the number of instances in the training set could increase in each iteration, it may allow to use larger neighborhoods, hence, in this variant the value of $k$ is increased after a predefined number of iterations.

\subsection{Pseudo-label an instance} \label{s:pseudo-label}
This section presents how pseudo-labels are built from the paths of labels of labeled instances.
Let $Y=[y_{1},...,y_{k}]$ be the labels of $k$ labeled instances close to the unlabeled instance $x$, where $y_{i} \in \{0,1\}^{ \mid L \mid}$ and each $y_{i,j}$ is 1 if the $i$-th instance is associated to the $j$-th label, 0 otherwise.  The probabilities for each individual label can be estimated as follows:
\begin{equation}
    ppsl_{j}= \frac{\sum_{i=1}^{k}y_{i,j}}{k}  , \: \forall j \in \{1,...,\mid L \mid\}
\end{equation}
Later, the threshold $t2label$  is used to determine whether an instance is associated to the label: 
\begin{equation}
    psl_{j} = \left\{\begin{matrix}
        1:\: & ppsl_{j}>=t2label \\ 
        0:\: & ppsl_{j}<t2label
    \end{matrix}\right., \: \forall j \in \{1,...,\mid L \mid\}
\end{equation}
\begin{equation}
    0<= t2label <= 1\nonumber
\end{equation}
In this way, $psl$ contains the pseudo label for $x$. Nevertheless, if $psl$ is full of \textit{zeros}, it means that $x$ did not get a valid pseudo-label and stays unlabeled.

\begin{figure}[tb]
	\centering
    \includegraphics[width=0.9\columnwidth]{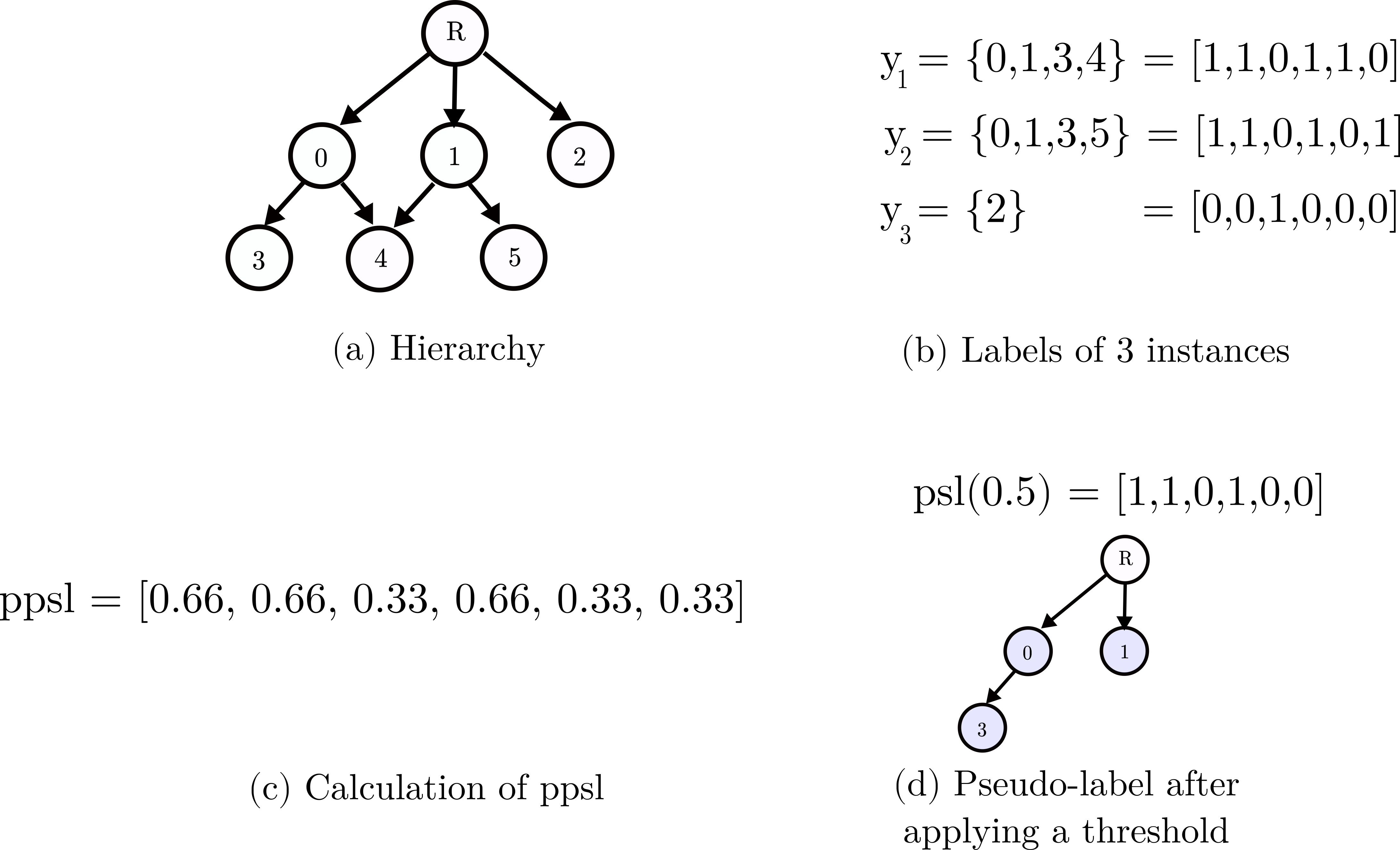}
    \caption{Example of how a pseudo-label is built. (b) shows the nodes, from the hierarchy (a), to which instances are associated, for example, the first instance is associated to   nodes $\{0,1,3,4\}$, next the vector form of this nodes is displayed. $ppsl$ is calculated with the labels in vector form of the instances in (b). Then the pseudo-label is obtained after applying a threshold to $ppsl$ as it is shown in (d).}
  \label{f:pl-mpl}
\end{figure}

Fig. \ref{f:pl-mpl} shows an example of how a pseudo-label is built from the paths of labels of 3 instances. The labels of the instances are required in vector form, which are  used to calculate $ppsl$, later, a threshold is applied to $ppsl$ which produces the pseudo-label as it is shown in \ref{f:pl-mpl}d. This way of pseudo-labeling has the advantage that can be applied to any hierarchy of tree or DAG type, furthermore, it can be applied to instances that are associated to multiple paths of labels, that is, it does not limit the pseudo-labels to be associated to a single path of labels.

\subsection{Base Classifier for SSHMC-BLI} \label{s:base-sshmc-bli}
In this case, a hierarchical classifier based on local classifiers per node (LCN) is trained, that is, for each node of the hierarchy (except the root node) a binary classifier is trained; for the experiments in this work,  \textit{random forest classifier}\footnote{Implementation of \url{scikit-learn.org}. Default parameters except \{$n\_jobs=5$\}. } is used as the local classifier, additionally, the policy \textit{balanced bottom-up} is used to select the positive and negative instances at each node. Later, in the prediction phase the probabilities of being associated to each LCN are obtained, then a post-processing is applied. The post-processing \citep{giunchiglia2020-CHMLCN} follows a top-down manner where the probabilities of each node are limited by the probabilities of their parents; let $a,b$ be nodes, let $f^{*}$ be the post-processed output of the model $f$ then:
\begin{equation}
    f^{*}_{a}=min_{b \in S_{a}}(f_{b}) \label{eq:fmin}
\end{equation}
\begin{equation*}
    S_{a} = Parents(a) \cup  \{a\} 
\end{equation*}
that is, a node gets the lowest probability among itself and its parents as shown in equation \ref{eq:fmin}.

In this way, the predictions of the model $f^{*}$ are consistent with the hierarchy, because they complain the \textit{hierachical probability constraint}, that is, the probability of every node is equal or lower than the probability of its parents.

\begin{figure}[ht]
	\centering
    \includegraphics[width=0.8\columnwidth]{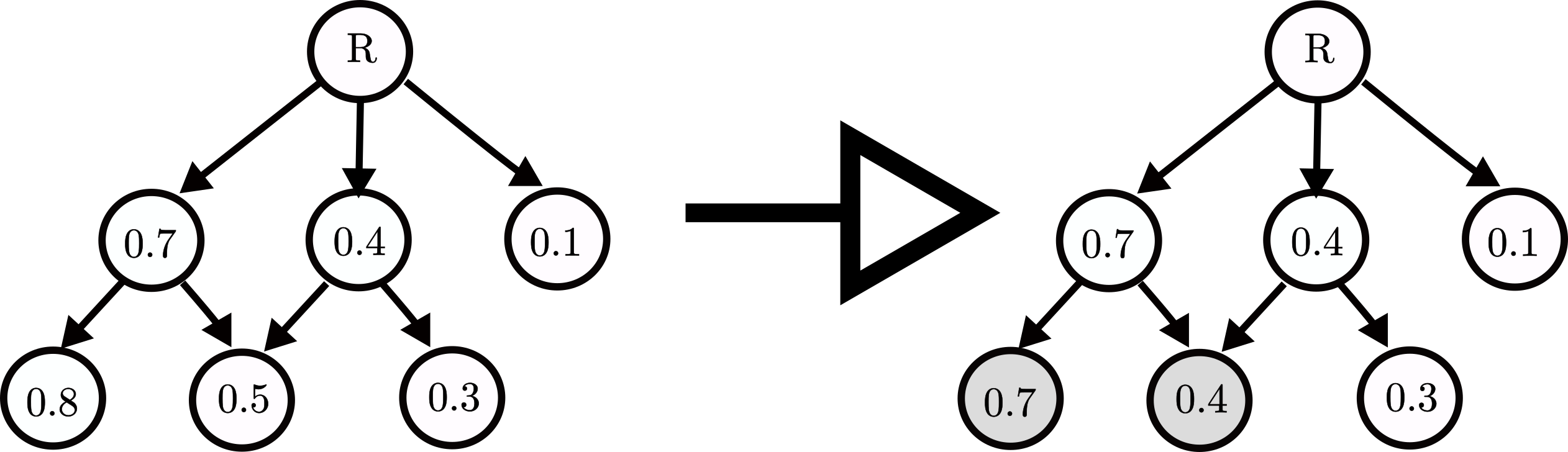}
    \caption{Example of post-processing. \textit{Left} hierarchy has the probability of being associated to each node. \textit{Right} hierarchy shows the post-processed output; grey nodes were limited by the probabilities of their parents.  } 
  \label{f:pp-mpl}
\end{figure}

An example of the post-processing is depicted in Fig. \ref{f:pp-mpl}. The left hierarchy shows the probabilities of being positively associated (before post-processing), those probabilities does not comply the \textit{hierarchical probability constraint}, so the post-processing is applied as it is shown in the hierarchy to the right, where the probabilities of a couple of nodes were limited in order to comply the hierarchical probability constraint.

\section{Datasets} \label{s:datasets}
Real world datasets from the field of functional genomics were collected, the datasets belong to the Gene Ontology (GO) collection \citep{vens2008decision}. The labels of each datasets are arranged in a hierarchy of directed acyclic graph type, that is, some nodes have more than one parent. Furthermore, the instances can be associated to multiple paths of labels which can finish in an internal node. 

The datasets are split into training, test and validation sets. However, they were preprocessed in similar fashion than  Ramírez-Corona et al. \cite{ramirez2016hierarchical}, that is, nodes with less than 50 instances associated in the training set were removed, then, the same nodes were removed from validation and test sets; however, in this case, all the paths to which instances are associated are kept. Description of GO datasets is shown in Table \ref{t:desc_go}; following the notation of Sillas and Freitas \cite{Silla2011}, those datasets are described as \textit{(DAG, MPL, PD)}.

\begin{table}[ht]
\centering
\caption{Description of Gene Ontology (GO) datasets. \textit{Train}, \textit{validation} and \textit{test} show the number of instances in each set; \textit{Attr.} shows the number of attributes; \textit{Nodes} shows the number of nodes/labels in the hierarchy; and  \textit{MD} correspond to the maximum depth of the hierarchy.}
\label{t:desc_go}
\begin{tabular}{@{}lcccccc@{}}
\toprule
\textbf{Dataset} & \textbf{Train} & \textbf{Validation} & \textbf{Test} & \textbf{Attr.} & \textbf{Nodes} & \textbf{MD} \\ \midrule
cellcycle\_GO    & 1625           & 848                 & 1278          & 77             & 164            & 9           \\
church\_GO       & 1627           & 844                 & 1278          & 31             & 164            & 9           \\
derisi\_GO       & 1605           & 842                 & 1272          & 63             & 161            & 9           \\
eisen\_GO        & 1055           & 528                 & 835           & 79             & 122            & 9           \\
expr\_GO         & 1636           & 849                 & 1288          & 565            & 165            & 9           \\
gasch1\_GO       & 1631           & 846                 & 1281          & 173            & 165            & 9           \\
gasch2\_GO       & 1636           & 849                 & 1288          & 52             & 165            & 9           \\
hom\_GO          & 1661           & 867                 & 1309          & 47034          & 166            & 9           \\
pheno\_GO        & 653            & 352                 & 581           & 276            & 68             & 7           \\
seq\_GO          & 1692           & 876                 & 1332          & 530            & 171            & 9           \\
spo\_GO          & 1597           & 837                 & 1263          & 89             & 162            & 9           \\
struc\_GO        & 1659           & 859                 & 1306          & 19628          & 169            & 9           \\ \bottomrule
\end{tabular}
\end{table}

\section{Experiments and Results}\label{s:ER}

The experiments are focused on showing that using unlabeled data may help to improve the performance of a hierarchical classifier trained only on labeled data.
The results of the proposed method are compared against standard methods and a supervised hierarchical (LCN) classifier, which can be seen as the base line, since it is only trained with labeled data

In order to carry out the experiments,  first, the training set of each dataset was split into labeled and unlabeled sets. Then, the best configuration of each method is obtained by varying their parameters, while they are evaluated on the validation set. Results of the methods, trained with their best configuration, on the test set are reported. Later, results of the variants of the proposed, standard and supervised methods were compared with the Friedman test to identify statistical difference among them.

\subsection{SSHMC-BLI Behaviour}
\begin{figure}[ht]
	\centering
	
    \begin{subfigure}{0.47\columnwidth}
        \includegraphics[width=\columnwidth]{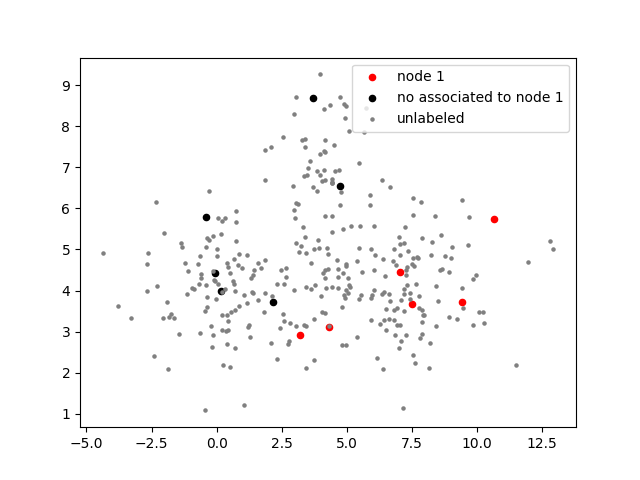}
        \caption{Node 1}
        \label{f:node_1}
    \end{subfigure}    
    \begin{subfigure}{0.47\columnwidth}
        \includegraphics[width=\columnwidth]{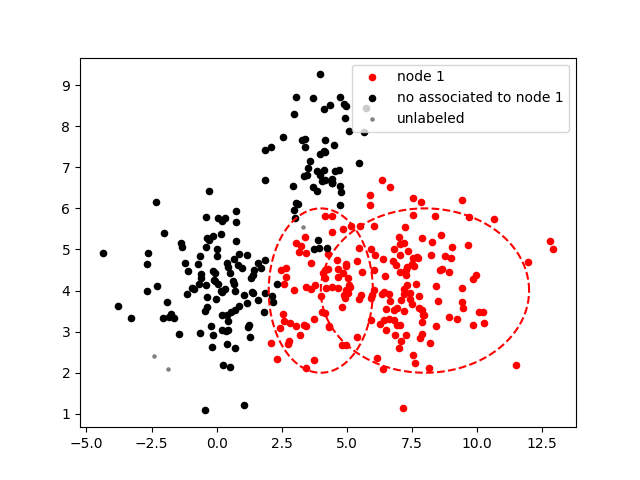}
        \caption{Node 1 - V1}
        \label{f:node_1_v1}
    \end{subfigure}
    
    \begin{subfigure}{0.47\columnwidth}
        \includegraphics[width=\columnwidth]{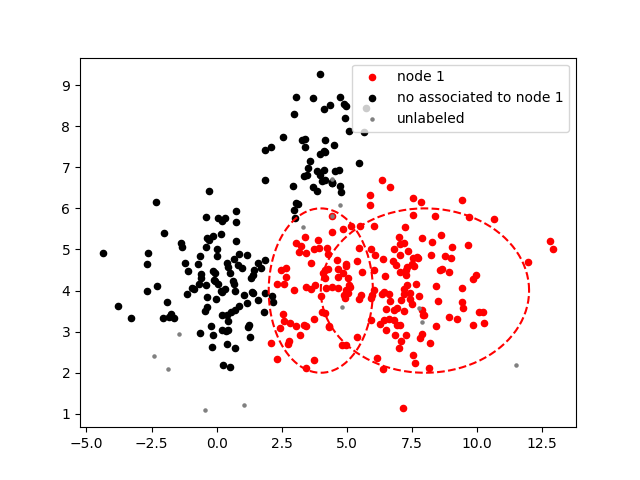}
        \caption{Node 1 - V2}
        \label{f:node_1_v2}
    \end{subfigure}    
    \begin{subfigure}{0.47\columnwidth}
        \includegraphics[width=\columnwidth]{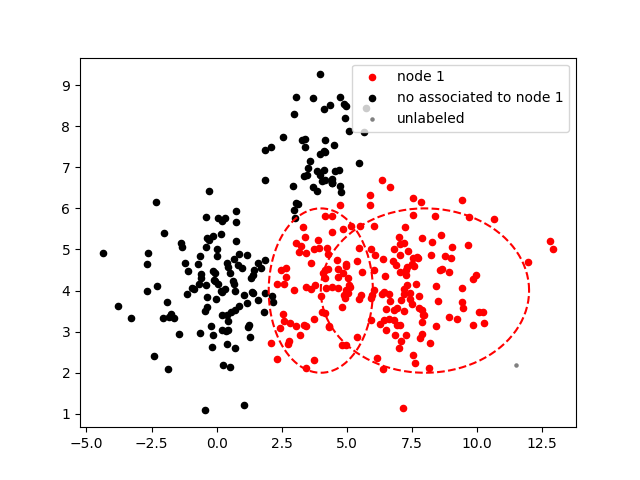}
        \caption{Node 1 - V3}
        \label{f:node_1_v3}
    \end{subfigure}    
    \caption{Pseudo-labels for the artificial dataset. (a) shows the initial data for node 1. (b,c,d) show the pseudo-labeled data at the end of each variant, V1, V2, V3, respectively. Inside the red circle is approximately 95\% of the data that should be associated to node 1.  (Best seen in color.) }
  \label{f:res-ad-mpl}
\end{figure}

\begin{figure}[ht]
	\centering
	
    \begin{subfigure}{0.47\columnwidth}
        \includegraphics[width=\columnwidth]{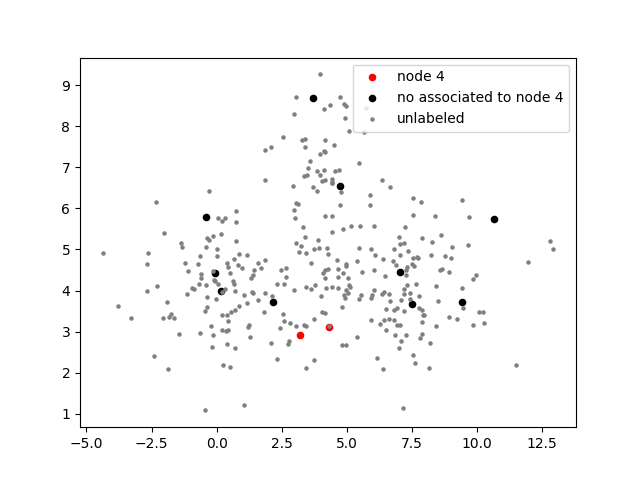}
        \caption{Node 4}
        \label{f:node_4}
    \end{subfigure}    
    \begin{subfigure}{0.47\columnwidth}
        \includegraphics[width=\columnwidth]{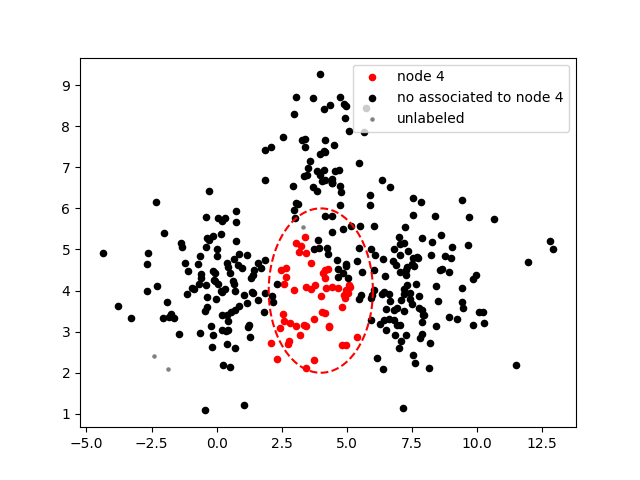}
        \caption{Node 4 - V1}
        \label{f:node_4_v1}
    \end{subfigure}
    
    \begin{subfigure}{0.47\columnwidth}
        \includegraphics[width=\columnwidth]{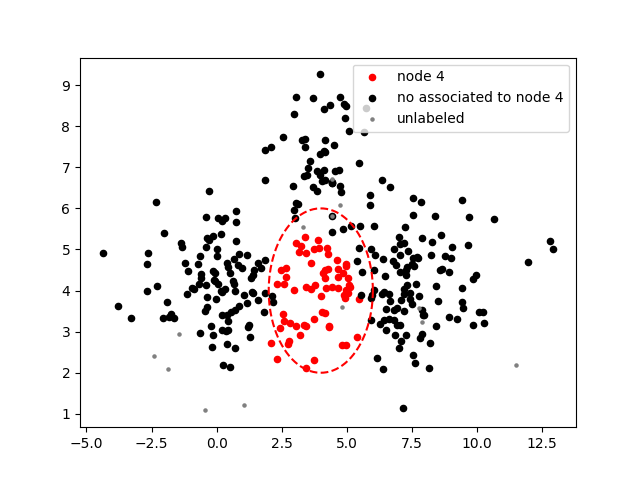}
        \caption{Node 4 - V2}
        \label{f:node_4_v2}
    \end{subfigure}    
    \begin{subfigure}{0.47\columnwidth}
        \includegraphics[width=\columnwidth]{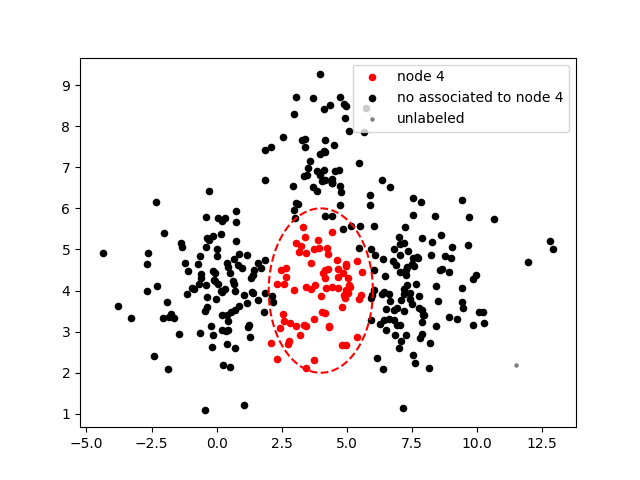}
        \caption{Node 4 - V3}
        \label{f:node_4_v3}
    \end{subfigure}
    \caption{Pseudo-labels for the artificial dataset. (a) shows the initial data for node 4. (b,c,d) show the pseudo-labeled data at the end of each variant, V1, V2, V3, respectively. Inside the red circle is approximately 95\% of the data that should be associated to node 4.  (Best seen in color.) }
  \label{f:res-ad-mpl_2}
\end{figure}

In order to show if the SSHMC-BLI variants pesudo-label properly the unlabeled data, an small artificial dataset was designed. The hierarchy used is shown in Fig. \ref{f:pl-mpl}a, which is composed by 6 nodes; the dataset is two-dimensional; 12 and 330 instances were generated from normal distributions for labeled and unlabeled datasets, respectively; plus 300 instances for test set. Figs. \ref{f:res-ad-mpl}a and \ref{f:res-ad-mpl_2}a show the instances associated to nodes 1 and 4 respectively, also the unlabeled instances are depicted.

The SSHC-BLI variants were applied to this dataset with the following configuration: nearest labeled neighbors, $k=3$; similitude threshold, $THR=0.5$; threshold to positively label an instance, $t2label=0.5$; for variant 3, $k$ increase each 10 iterations. Finally, a LCN (section \ref{s:base-sshmc-bli}) classifier was trained with labeled and pseudo-labeled data as the base classifier of the method.

Figs. \ref{f:res-ad-mpl} and \ref{f:res-ad-mpl_2} show the way of how SSHMC-BLI variants pseudo-labeled the unlabeled data for a couple of nodes, 1 and 4, respectively; inside the red circles is approximately 95\% of the data that should be associated to the corresponding node. Variant 1 pseudo-labeled almost the whole unlabeled set, but it wrongly pseudo-labeled some instances as it can be seen in Fig. \ref{f:res-ad-mpl_2}b; however, most of them are properly pseudo-labeled by its parent node as it can be seen in Fig. \ref{f:res-ad-mpl}b.
Variant 2 pseudo-labeled in a better way the instances associated to nodes 1 and 4, as it can be seen in Figs. \ref{f:res-ad-mpl}c and \ref{f:res-ad-mpl_2}c; however, it was the variant where more instances stayed unlabeled. 
Furthermore, some unlabeled instances are surrounded by labeled instances with the same labels, as it can be seen in Fig. \ref{f:res-ad-mpl}c, so one could think that they must have the same labels, nevertheless, they were no pseudo-labeled due to the estimation of the similitude with its nearest neighbors.
Finally, variant 3 seems to smooth the results obtained by 2nd variant, see Figs. \ref{f:res-ad-mpl}d and \ref{f:res-ad-mpl_2}d, since it was able to pseudo-label most of the instances that were previously not pseudo-labeled.

Later, for each variant the labeled and pseudo-labeled instances are used to train a hierarchical classifier. Results obtained by the SSHMC-BLI variants are shown in Table \ref{t:sshc-knn-ha02}; in the same table, results obtained by the supervised classifier (LCN), trained only with labeled data, are reported. As it can be seen, the results obtained by the SSHMC-BLI variants outperformed the result of the supervised classifier. 

\begin{table}[ht]
\centering
\caption{Results of SSHMC-BLI variants (1, 2 and 3) and the supervised classifier, LCN, for the artificial dataset. In bold the best score.}
\label{t:sshc-knn-ha02}
\begin{tabular}{@{}lcccc@{}}
\toprule
                       \textbf{Measure} & \textbf{LCN} & \textbf{V1}     & \textbf{V2} & \textbf{V3} \\ \midrule
Avg. precision & 0.4492       & \textbf{0.4817} & 0.4526      & 0.4573      \\ \bottomrule
\end{tabular}
\end{table}

\subsection{GO datasets}
\begin{figure}[h!]
	\centering
	
    \begin{subfigure}{0.45\columnwidth}
        \includegraphics[width=\columnwidth]{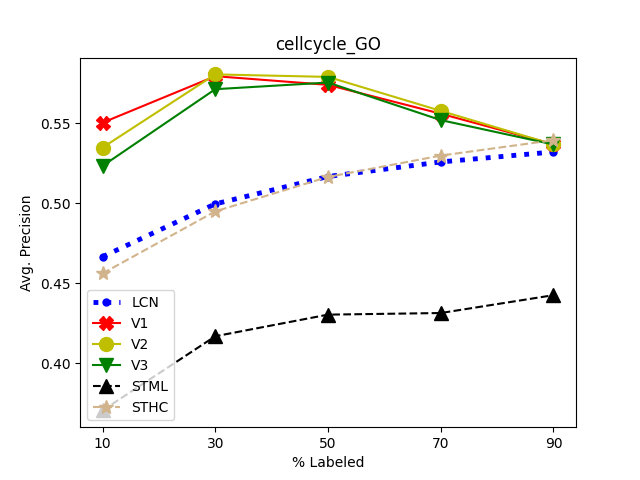}
        \caption{}
        \label{f:go_cellcycle}
    \end{subfigure}
    \begin{subfigure}{0.45\columnwidth}
        \includegraphics[width=\columnwidth]{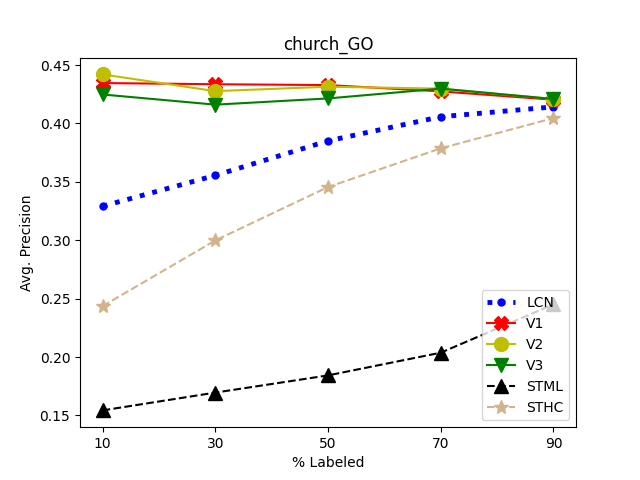}
        \caption{}
        \label{f:go_church}
    \end{subfigure}
    
    \begin{subfigure}{0.45\columnwidth}
        \includegraphics[width=\columnwidth]{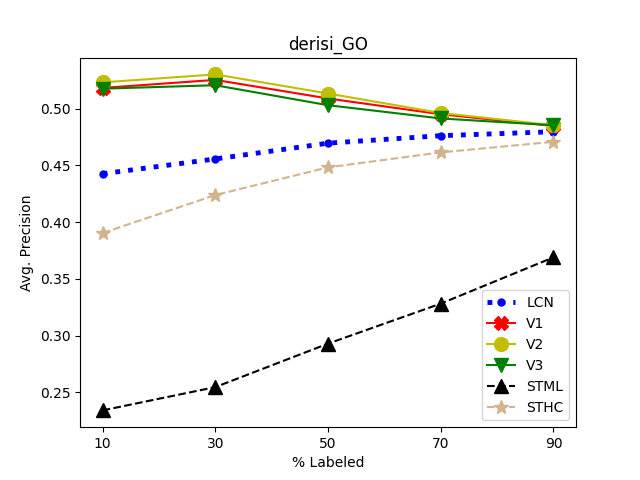}
        \caption{}
        \label{f:go_derisi}
    \end{subfigure}    
    \begin{subfigure}{0.45\columnwidth}
        \includegraphics[width=\columnwidth]{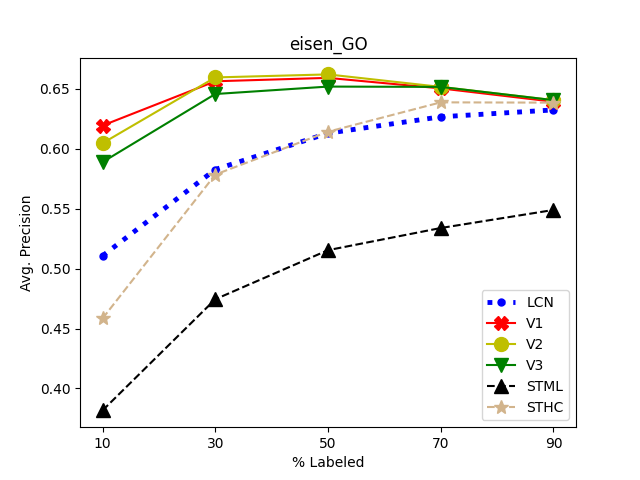}
        \caption{}
        \label{f:go_eisen}
    \end{subfigure} 
    
    \begin{subfigure}{0.45\columnwidth}
        \includegraphics[width=\columnwidth]{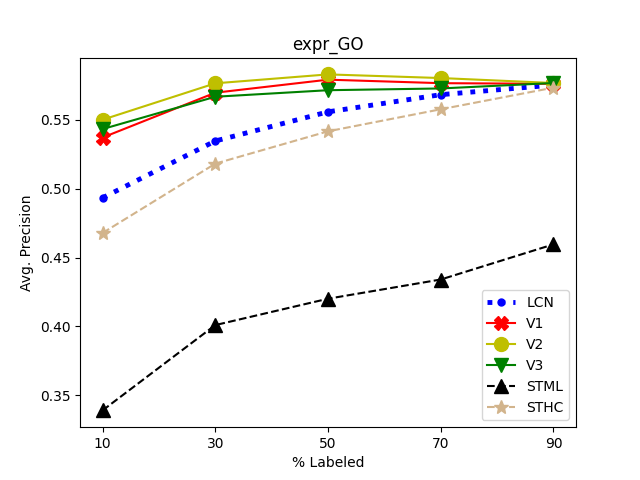}
        \caption{}
        \label{f:go_expr}
    \end{subfigure} 
    \begin{subfigure}{0.45\columnwidth}
        \includegraphics[width=\columnwidth]{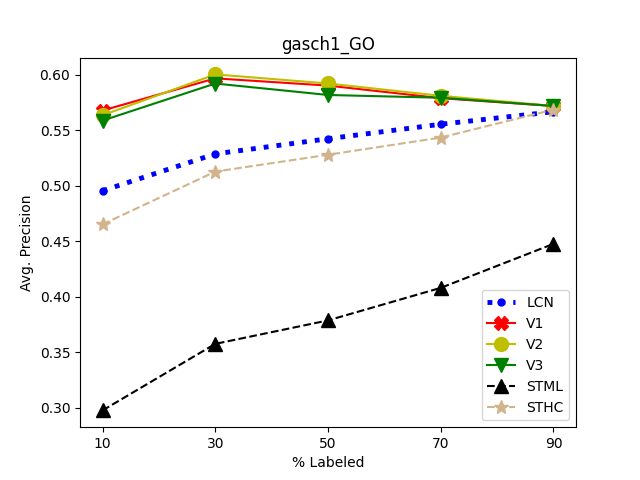}
        \caption{}
        \label{f:go_gasch1}
    \end{subfigure}
        
    \caption{Results in different GO datasets with the evaluation measure average precision: a) cellcycle, b) church, c) derisi, d) eisen, e) expr and f) gasch1. The $x$-axis correspond to the amount of labeled data (while its complement is the unlabeled data). (Best seen in color.)}
  \label{f:res-go}
\end{figure}

\begin{figure}[h!]
	\centering	    
    \begin{subfigure}{0.45\columnwidth}
        \includegraphics[width=\columnwidth]{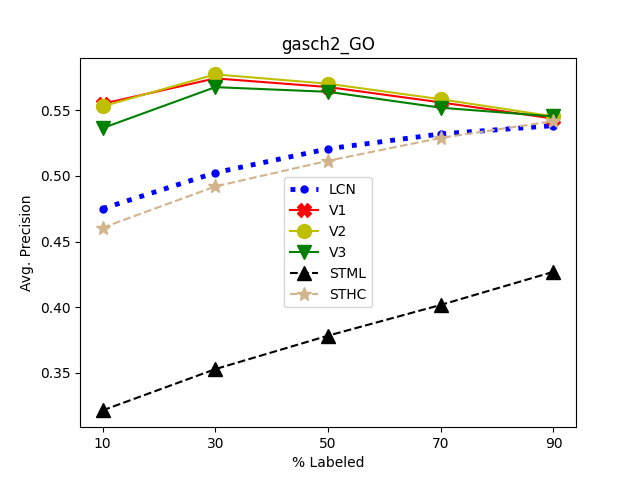}
        \caption{}
        \label{f:go_gasch2}
    \end{subfigure}
    \begin{subfigure}{0.45\columnwidth}
        \includegraphics[width=\columnwidth]{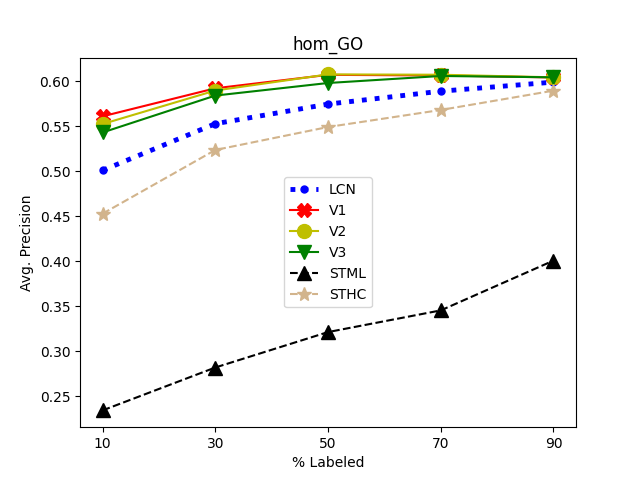}
        \caption{}
        \label{f:go_hom}
    \end{subfigure}
    
    \begin{subfigure}{0.45\columnwidth}
        \includegraphics[width=\columnwidth]{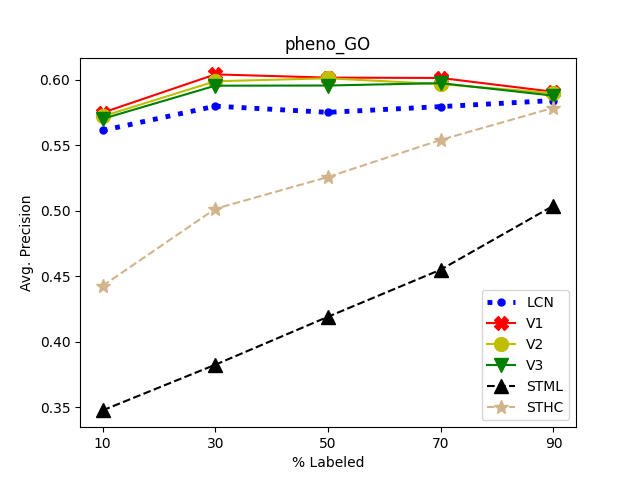}
        \caption{}
        \label{f:go_pheno}
    \end{subfigure}
    \begin{subfigure}{0.45\columnwidth}
        \includegraphics[width=\columnwidth]{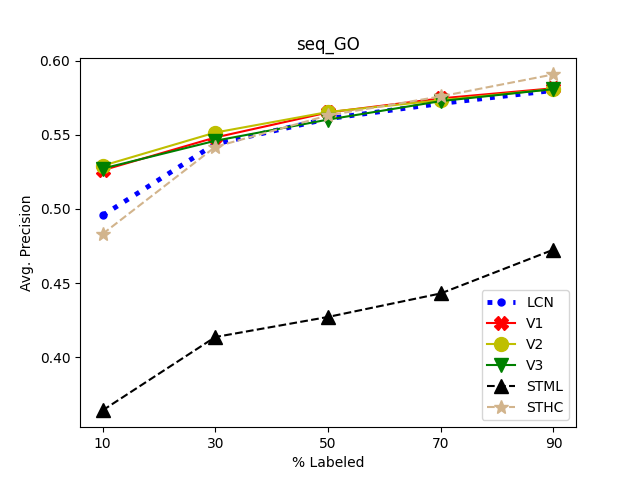}
        \caption{}
        \label{f:go_seq}
    \end{subfigure}
    
    \begin{subfigure}{0.45\columnwidth}
        \includegraphics[width=\columnwidth]{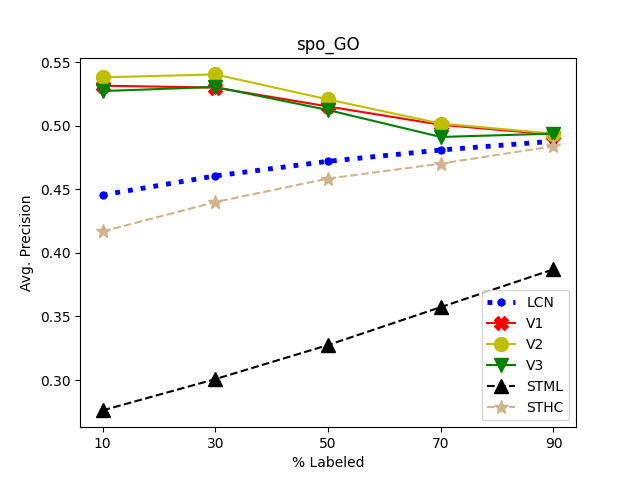}
        \caption{}
        \label{f:go_spo}
    \end{subfigure}
    \begin{subfigure}{0.45\columnwidth}
        \includegraphics[width=\columnwidth]{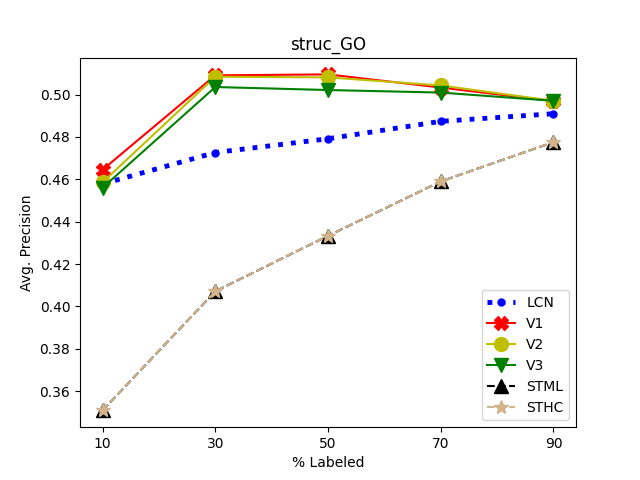}
        \caption{}
        \label{f:go_struc}
    \end{subfigure}
    \caption{Results in different GO datasets with the evaluation measure average precision: a) gasch2, b) hom, c) pheno, d) seq, e) spo and f) struc. The $x$-axis correspond to the amount of labeled data (while its complement is the unlabeled data). (Best seen in color.)}
  \label{f:res-go2}
\end{figure}

The GO datasets are already divided into training and test sets. However, the training sets were stratified split in labeled and unlabeled sets as follows: 
\begin{itemize}
    \item Labeled: \{10, 30, 50, 70, 90\}\%
    \item Unlabeled: \{90, 70, 50, 30, 10\}\%, complement with respect to labeled.
\end{itemize}
Division of training set was carried out 3 times, so results are the averages. 

The validation set was used for tuning the hyper-parameters of the SSHMC-BLI for each run. The parameters and values that could take are the following: similitude threshold (THR): \{0.3, 0.5, 0.7\}; number of labeled neighbors (k): \{3, 4, 5\}.
The evaluation measure AP was used to determine the best configuration. Later, the semi-supervised classifier is trained with the best configuration but only with the labeled and unlabeled sets; that is, the validation set is not used for training.  Finally, the SSHMC-BLI classifier is evaluated in the test set.

The results in average precision of the SSHMC-BLI variants, the standard methods (STML and STHC) and the supervised (LCN) classifier are shown in Figures \ref{f:res-go} and \ref{f:res-go2}. STML got the worse performance, since it is a multi-label method, it do not consider the hierarchy when training and the predictions do not always comply the hierarchical probability constraint. STHC improved the performance of STML just by adding the post-processing to comply the hierarchical probability constraint, nevertheless, its performance is most of times lower than the supervised, showing that the self-training strategy does not help to improve the performance of the supervised classifier. Finally, the different variants of the SSHMC-BLI got the best performances among the different semi-supervised and the supervised methods. The greatest improvement is found when there is just few labeled instances, then the performance of the SSHMC-BLI variants became closer to the supervised as the amount of labeled data increase.

\subsection{Discussion and Statistical Analysis}
Table \ref{t:avg-ranks-go} summarizes the results of the semi-supervised variants and the supervised classifier, that is, it presents the average rank of each classifier in the datasets, where the SSHC-BLI variants obtained the best performance. Variant 2 got the best performance, this variant does not allow to pseudo-labeled instances to make use of themselves when building the new pseudo-labels; situation that was beneficial in these datasets.

\begin{table}[ht]
\centering
\caption{Average rank of each classifier in the GO collection. In bold the best (lower is better).}
\label{t:avg-ranks-go}
\begin{tabular}{@{}lcccccc@{}}
\toprule
               & \textbf{LCN} & \textbf{V1} & \textbf{V2} & \textbf{V3} & \textbf{STML} & \textbf{STHC} \\ \midrule
Avg. Precision  & 4.133 & 2.0 & \textbf{1.492} & 2.708 & 5.958 & 4.708 \\ \bottomrule
\end{tabular}
\end{table}


The SSHMC-BLI variants got their best performance when there is few labeled data in most of the GO datasets, then their performance tend to decrease as the amount of labeled data increase, becoming closer to the performance of the supervised classifier. We attribute this behavior to the  fact that those datasets are very hard and noisy. Moreover, Vens et al. \cite{vens2008decision} indicate that some GO datasets suffer from non-unique feature representations, situation that could also affect the performance of the semi-supervised classifiers, likewise in Pliakos et al. \cite{Pliakos-RPGFFP-2015}. 
In other words, we hypothesize that when the training set is divided into labeled and unlabeled sets, some noisy data is removed from the labeled set, now in unlabeled set; so, as the amount of labeled data is increased, also the amount of noisy data, situation that would explain the decrease in performance when the amount of labeled data is increased.

On the other hand, in order to estimate if there is statistical difference among the SSHMC-BLI variants, standard methods and the supervised classifier, the Friedman test together with its post-hoc the Nemenyi test were used, as recommended by  Demsar \cite{Demsar:2006:SCC:1248547.1248548} when comparing multiple classifiers over multiple datasets.

First, let $r_{i}^{j}$ be the rank of the $j$-th of $l$ algorithms on the $i$-th of M datasets, then $R_{j}=\frac{1}{M} \sum_{i=1}^{M}{r_{i}^{j}}$ is the average rank of the $j$-th algorithm. So, the null hypothesis of the Friedman test states that all the algorithms are equivalent, therefore their average ranks ($R_{j}$) should be equal, against the alternative which states that they are not. 
Afterward, only if the null hypothesis was rejected, the Nemenyi test is used to compare all the classifiers against each other. Hence, the performance of two classifiers is significantly different if their average ranks differ by at least the \textit{critical difference}.

\begin{figure}[ht]
	\centering    
    \includegraphics[width=0.9\columnwidth]{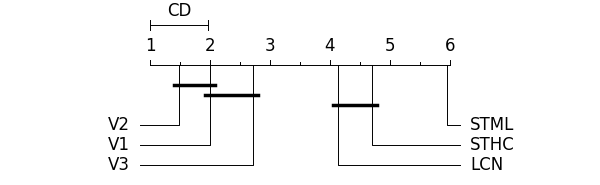}    
    \caption{Graphical representations of Nemenyi test for the evaluation measures average precision. Classifiers that are not significantly different, with $p=0.05$, are connected. CD: critical difference. (Lower is better)}
  \label{f:nemenyi_go}
\end{figure}

The datasets of the Gene Ontology collection (and their corresponding divisions each) were considered. The result of the Friedman test with $p=0.05$ for evaluation measure \textit{average precision} is that the null hypothesis can be rejected in favor of the alternative, that is, the average ranks of the algorithms are not equal.
Since the null hypothesis was rejected, now we can proceed with the Nemenyi tests. Fig. \ref{f:nemenyi_go} shows the graphical representation of the Nemenyi test for average precision. The three variants of SSHMC-BLI are significantly different than STML, STHC and the supervised method, LCN.

Throughout the experiments in the different datasets, the SSHMC-BLI variants obtained the best performances. Variant 2 got the best performance among the variants in general, although for some datasets the variant 1 is sightly better. However, the Nemenyi test indicates that the performances of variants 1 and 2 are not significantly different.

Finally, it is expected that the SSHMC-BLI variants work properly on scenarios were close instances share the same or similar paths of labels, that is, when the datasets fulfill the smoothness assumption. On the other hand, the SSHMC-BLI will have its performance limited when the datasets do not fulfill the smoothness assumption, that is, when close instances do not necessarily share the same or similar path of labels. 

\section{Conclusions and Future Work}\label{s:CFW}

In this manuscript the semi-supervised hierarchical multi-label classifier based on local information (SSHMC-BLI) was proposed (available as open source). This method is based on the \textit{smoothness assumption}, which tries to pseudo-label the unlabeled instances making use of the paths of labels of their labeled neighbors, while considering if the unlabeled instances are similar to their labeled neighbors. The method is general enough to be applied to any hierarchical problems, since it can handle any hierarchy of directed acyclic graph type, the instances can be associated to multiple paths of labels, and the paths that can finish in internal nodes.
The method was evaluated in the most difficult hierarchical problem: hierarchies of DAG type, where the instances can be associated to multiple paths of labels which can finish in an internal node. From the best of our knowledge, SSHMC-BLI is the first proposed method to be applied in the aforementioned hierarchical problem.




Experiments on the Gene Ontology datasets were carried on, where it was shown that making use of unlabeled data along with labeled, in the aforementioned hierarchical classification scenario, can help to get a semi-supervised hierarchical classifier with better performance than a hierarchical classifier trained only on the labeled data. 
Furthermore, the Friedman test along with its post-hoc the Nemenyi test were applied to the results of the methods, showing that the performance of the proposed method is significantly better than the obtained by the standard (STML, STHC) and the supervised methods.


As future work, the proposed method will be combined with deep neural networks so it could be applied to image classification problems where the labels are arranged in a hierarchy.


\section*{Acknowledgments}
J. Serrano-Pérez acknowledges the support from CONAHCYT scholarship number (CVU) 84075.




\bibliographystyle{elsarticle-num-names} 
\bibliography{mybibliography}





\end{document}